\documentclass[letterpaper,10pt,conference]{ieeeconf}
\usepackage[utf8]{inputenc}
\usepackage{listings}
\usepackage{graphicx}
\usepackage{caption}
\usepackage{subfig}
\usepackage{url}
\usepackage{color}
\usepackage{citesort}

\IEEEoverridecommandlockouts 
\lstdefinelanguage{dolphin}{
  morekeywords={
poll,action,condition,post,location,def,test,consume,
pick,message,pause,idle,message,task,execute,waitFor,
select,then,allOf,oneOf,ask,setConnectionTimeout,
choose,when,run,area,release,
halt,ignore,propagate,during,until,position,watch,onError,while},
comment=[l]{//},
morecomment=[s]{/*}{*/},
morestring=[b]",
basicstyle={\small\bfseries\ttfamily},
keywordstyle={\small\bfseries\color{blue}\ttfamily},
numberstyle={\scriptsize\ttfamily},
aboveskip=0pt,
belowskip=0pt,
boxpos=t,
escapechar=@
}
\title{Dolphin: a task orchestration language for autonomous vehicle networks}

\author{Keila Lima$^{1}$, Eduardo R. B. Marques$^{2}$, José Pinto$^{1}$, João B. Sousa$^{1}$%
\thanks{
$^{1}$ Laboratório de Sistemas e Tecnologia Subaquática, Faculdade de Engenharia da Universidade do Porto, Portugal.
}
\thanks{
$^{2}$ CRACS/INESC-TEC \& Faculdade de Ciências da Universidade 
do Porto, Portugal.
}
}

\begin{document}

\maketitle
\thispagestyle{empty}
\pagestyle{empty}

\newcommand{\psection}[1] {\vspace{0.1cm}\noindent{\bf #1.}}

\bstctlcite{IEEEexample:BSTcontrol}

\begin{abstract}
We present Dolphin, an extensible programming language for autonomous vehicle networks. 
A Dolphin program expresses an orchestrated execution of tasks defined compositionally for multiple vehicles. 
Building upon the base case of elementary one-vehicle tasks, the built-in operators include support for composing tasks in several forms, for instance according to concurrent, sequential, or event-based task flow. 
The language is implemented as a Groovy DSL, facilitating extension and integration with 
external software packages, in particular robotic toolkits. The paper describes the Dolphin 
language,  its integration with an open-source toolchain for
autonomous vehicles, and results from field tests using unmanned underwater vehicles (UUVs) and unmanned
aerial vehicles (UAVs).
\end{abstract}
\vspace{-0.1cm}\section{Introduction}

The use of autonomous vehicles is now mainstream for several applications, in
particular those making use of several vehicles deployed at once for a common
purpose, in networked integration with sensors, human users, and
cyber-infrastructures~\cite{dunbabin,Wynn,sunrise,nsf-ooi}.
As part of these developments, several software toolkits became popular for
networked operation of autonomous vehicles~\cite{ros,mavlink,oceans13}, allowing
for remote control of a single vehicle or basic forms of networked interaction among vehicles.
To program a network of autonomous vehicles as an integrated whole, though, 
we feel that high-level abstractions are required, 
materialised by the use of domain-specific languages (DSLs) that
directly capture the modelling traits of multi-vehicle applications.

In particular, we are concerned with mixed-initiative systems, where 
humans are part of the control loop and burdened
by the intricate complexity of a system-of-systems~\cite{mixed-initiative}.
In multi-vehicle applications, part of this burden results from the need of 
separately programming each vehicle without principled mechanisms
for coordinated behavior, and the lack of a convenient abstraction 
for the global state of the system once it is deployed. 
To attack these problems, we advocate that humans-in-the-loop should be able to write
programs that orchestrate the a global behavior of multiple vehicles.

This motivation led us to the development of Dolphin, an extensible
task orchestration language for autonomous vehicle networks, that is available
open-source~\cite{dolphin-site}.
A Dolphin program expresses an orchestrated execution of tasks 
defined compositionally for multiple vehicles dynamically available in a network.
Building upon the base case of elementary one-vehicle tasks, the built-in operators
include support for composing tasks in several forms, for instance according to
concurrent, sequential, or event-based operators, partially inspired by process calculi approaches,
e.g., Milner's CCS~\cite{ccs}. The core language is agnostic
and independent of the underlying platform for networked vehicle operations. The
system is concretely instantiated through the implementation of abstract
programming bindings at the platform level of a robotic toolkit.
This is facilitated by the design of Dolphin as a Groovy domain-specific
language (DSL)~\cite{groovyDSL}, allowing direct integration/embedding with/in other
Groovy/Java software packages, and seamless addition of extended DSL features.

We developed Dolphin bindings for an open-source 
toolchain~\cite{oceans13} developed by Laboratório de Sistemas e Tecnologia Subaquática (LSTS)\footnote{\url{http://github.com/LSTS}},
used to operate heterogeneous types of unmanned vehicles in several experiments over the years (e.g.,~\cite{molamola,Chasing-Fish,icra14}), and bindings for the MAVLink drone
protocol~\cite{mavlink} are in progress~\cite{dolphin-site}.
The LSTS toolchain includes IMC, a message-based interoperability protocol, that
has a dedicated subset for the specification and execution of single-vehicle
tasks, called IMC plans. Using Dolphin, we were able to orchestrate IMC plans in expressive
manner for multiple vehicles in field tests using unmanned underwater vehicles
(UUVs) and unmanned aerial vehicles (UAVs). We present a field test scenario
where three UUVs concurrently perform a bathymetry survey over a given area,
and one simulated UAV also in the control loop engaged in a
rendezvous maneuver with each of the UUVs at a time.

The rest of the paper is structured as follows.
In section~\ref{sec:dolphin} we present Dolphin in terms of the underlying
architecture, an example scenario, task definition operators,
execution engine, and platform bindings.
Section~\ref{sec:lsts} describes the integration of Dolphin with the LSTS
toolchain and the related tools developed for that purpose.
Section~\ref{sec:experiments} reports results of field test experiments
we conducted for the example scenario using multiple autonomous vehicles. 
Section~\ref{sec:rwork} discusses related work. Finally, Section~\ref{sec:conclusion}
ends the paper with concluding remarks and a discussion of future work.

\section{The Dolphin language}
\label{sec:dolphin}

\subsection{Architecture}

\begin{figure}[t]
\centering
\includegraphics[width=0.8\columnwidth]{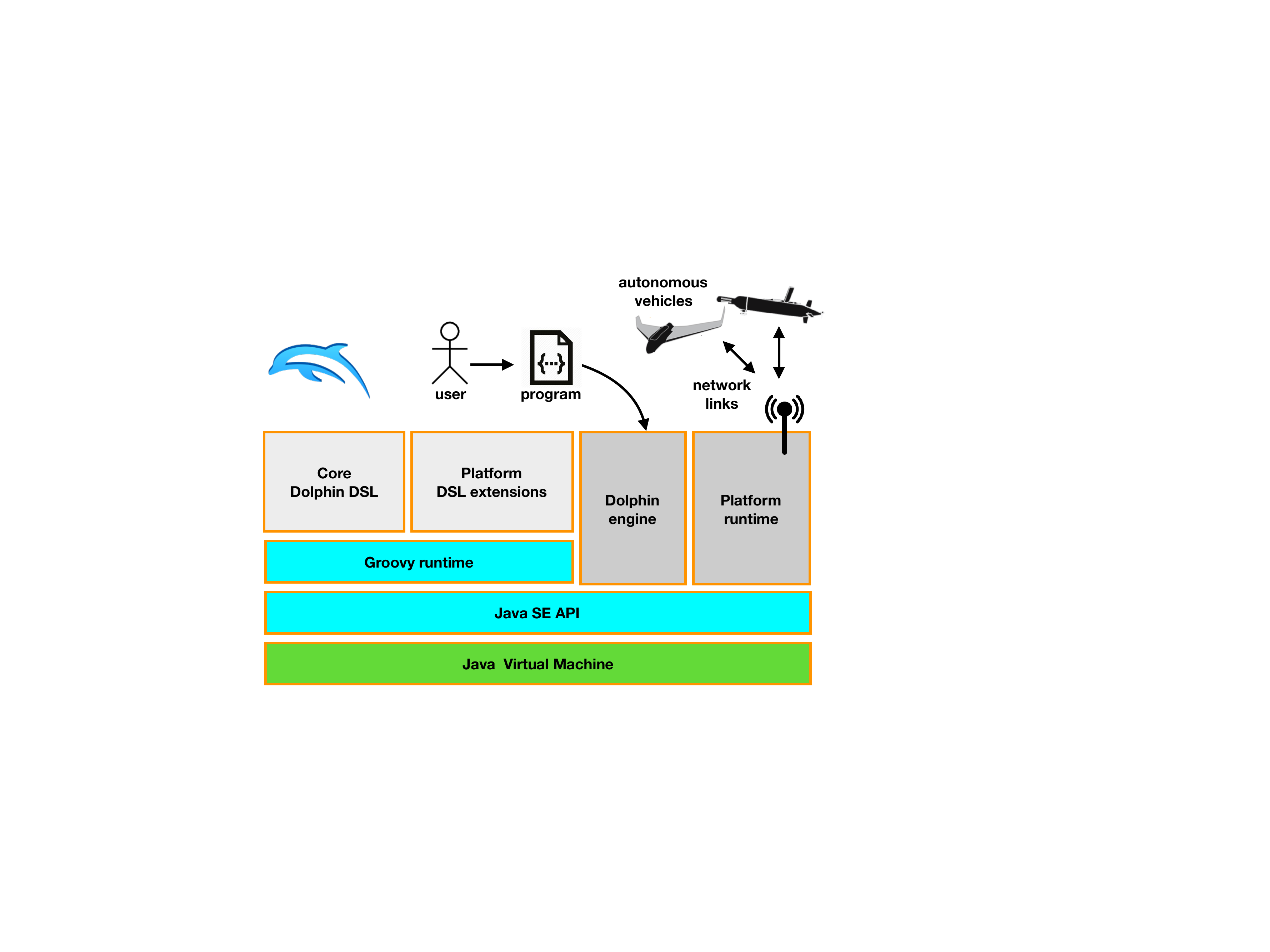}
\caption{Dolphin architecture.\label{fig:arch}}
\vspace{-0.4cm}
\end{figure}

The architecture of Dolphin is illustrated in Fig.~\ref{fig:arch}.
The language engine takes a program supplied by the user and executes it,
delegating platform-dependent networked operations to the platform runtime,
e.g., polling vehicles in the network or firing tasks for vehicles.
Thus, the execution of a Dolphin program is centralised, in interface with
networked vehicles. We believe this is a convenient approach for
mixed-initiative systems, allowing a human to orchestrate an entire 
network of vehicles based on a global specification.
%Regarding the communication between the language runtime and the nodes, there is support to volatile connections through the \lstinline|setConnectionTimeout| instruction. This instruction allows to define an amount of time during which the state of the vehicles is assumed to be fulfilling the program, originating an error in case the timeout is reached.
Furthermore, it does not
assume the need for peer-to-peer communication among vehicles or any form of
tightly coupled interaction among them, even if, of course, these aspects may be
relevant for many applications.
Later in the paper, we contrast this approach with others (Sec.~\ref{sec:rwork})
and identify aspects of future work to handle some of the inherent limitations
(Sec.~\ref{sec:conclusion}).

In the Dolphin architecture, the base components support the core 
DSL embedded in Groovy, and the engine required to execute programs in Java.
Platform instantiations may extend the DSL, and must provide a runtime for
networked interaction, implementing abstract Java bindings provided by the
Dolphin engine. The platform's DSL extensions and runtime are responsible
for implementing suitable constructs for platform tasks and their implementation.
The use of Groovy provides a number of
features useful for defining the DSL, e.g., operator overloading, meta-class
programming, or the use of closures~\cite{groovyDSL}.
Moreover, Groovy is fully interoperable with the Java
SE API and the Java Virtual Machine.
%this means Dolphin, Groovy and Java code is compiled
%onto JVM bytecode for execution.

\subsection{Example program}\label{sec:dolphin:example}

\begin{figure}[t!]
\centering
\includegraphics[width=0.7\columnwidth]{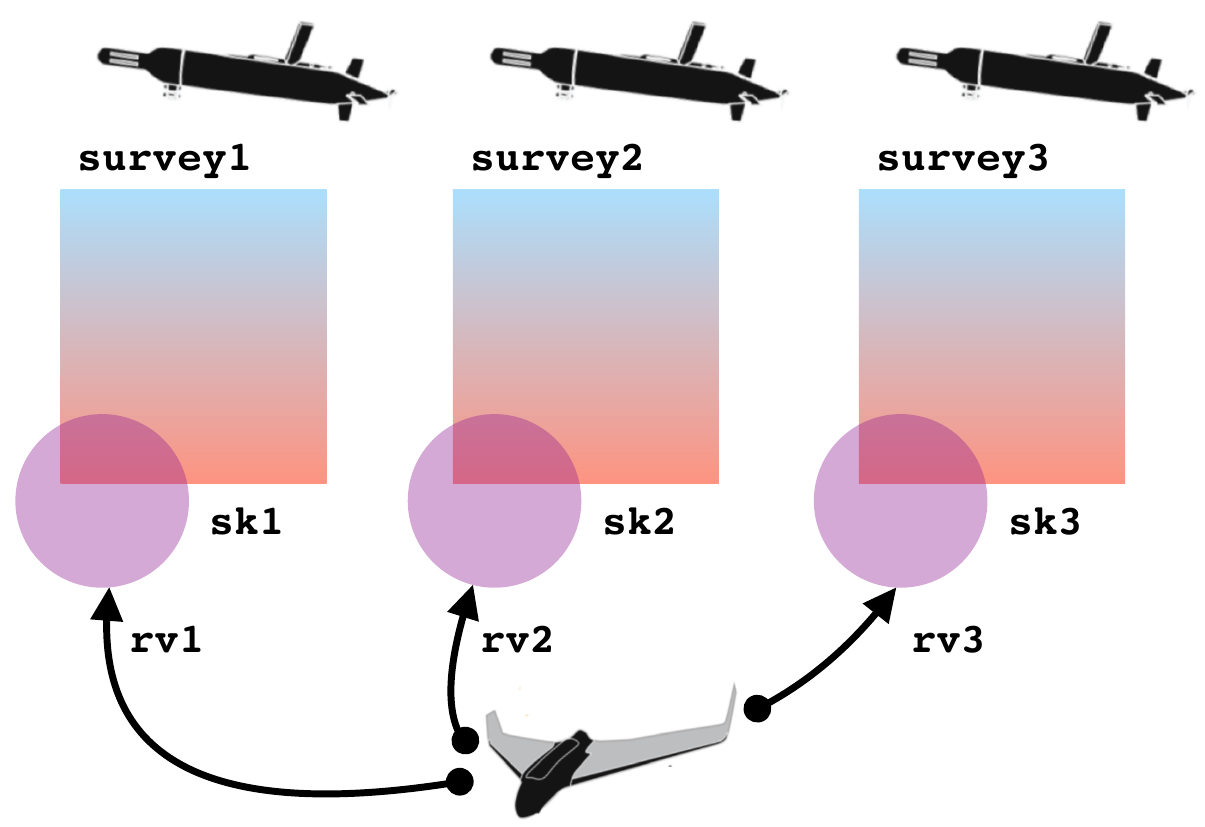}
\caption{Example scenario.\label{fig:example-scenario}}
\end{figure}

We now present an example scenario and a corresponding Dolphin program.
The scenario at stake, a generalisation of an example given in~\cite{sac15}, is
illustrated schematically in Fig.~\ref{fig:example-scenario}.
It comprises the use of 3 UUVs for joint surveys, executing concurrently over a
given area, and of a UAV that responds to the completion of each individual UUV
survey by approaching that UUV with a rendezvous behavior (e.g., to
retrieve survey data on-the-fly).
In Section~\ref{sec:experiments} we present an actual configuration and deployment
of a variant of this scenario in field tests; here we merely concentrate on its
overall meaning and realisation by a Dolphin program.

\begin{figure}[h!]
{\begin{lstlisting}[xleftmargin=6em,language=dolphin,basicstyle=\scriptsize\bfseries\ttfamily, 
commentstyle=\color{red}\bfseries,numberstyle=\scriptsize\ttfamily,keywordstyle=\scriptsize\bfseries\color{blue}\ttfamily,numbers=left]
// (1) Configuration @\label{c:conf:beg}@
r = ask 'Radius of operation area? (km)'
APDL = (location 41.18500, -8.70620) ^ r.km @\label{c:conf:end}@
// (2) Vehicle selection @\label{c:sel:beg}@
UUVs = pick {  
         count 3 @\label{c:conf:pick:UUVs}@
         type 'UUV'
         payload 'DVL','Sidescan'
         region APDL
       }
UAV  = pick { @\label{c:conf:pick:UAV}@
         type 'UAV'
         region APDL
       } @\label{c:sel:end}@
setConnectionTimeout UUVs, 2.min @\label{c:conf:timeout}@
// (3) Function yielding UUV task i @\label{c:aux:beg}@
def UUVTask ( i ) {
  imcPlan('survey' + i) >>
    action { post ready:i } >>
      imcPlan {
        planName 'sk' + i
        skeeping  duration: 600
      }
} @\label{c:aux:end}@
// (4) Execute tasks  @\label{c:exec:beg}@
execute UUVs: 
          UUVTask(1) | 
          UUVTask(2) | 
          UUVTask(3) ,
        UAV:
          allOf { 
            when { consume ready:1 } 
              then imcPlan('rv1')
            when { consume ready:2 } 
              then imcPlan('rv2')
            when { consume ready:3 } 
              then imcPlan('rv3')
          } @\label{c:exec:end}@
// (5) End @\label{c:end:beg}@
release UUVs + UAV @\label{c:release}@
message 'Done!' @\label{c:end:end}@
\end{lstlisting}
}
\caption{Dolphin program for the example scenario.\label{fig:program}}
\vspace{-0.4cm}
\end{figure}

The Dolphin program is listed in Fig.~\ref{fig:program}.
The code is basically structured in 5 segments:
(1) configuration (lines~\ref{c:conf:beg}--\ref{c:conf:end}); (2) vehicle selection (\ref{c:sel:beg}--\ref{c:sel:end}); (3)  an auxiliary function for 
parameterising the UUV tasks (\ref{c:aux:beg}--\ref{c:aux:end}); (4) the execution of desired tasks
using all 4 vehicles (\ref{c:exec:beg}--\ref{c:exec:end}), and; (5) program termination, decoupling
the vehicles from the program and displaying  a simple final message (\ref{c:end:beg}--\ref{c:end:end}).

The program begins by asking the user to input a radius \lstinline{r} of a geo-referenced area named \lstinline{APDL}. The UUVs and UAVs are then selected through two
\lstinline{pick} blocks, one for the UUVs and another one for the UUV.
The requirements are that the vehicles are located within the bounds of \lstinline{APDL}, and,
additionally, that each UUV is equipped with specific payload components, a Doppler velocity logger (DVL), 
and a side-scan sonar. Both the \lstinline{UUVs} and \lstinline{UAV} variables stand for vehicle sets 
(\lstinline{UAV} is a singleton) that can later be bound to the execution of tasks.
Vehicle sets can be manipulated using standard set operators, e.g., \lstinline{a + b} (as in line~\ref{c:release} of the program), \lstinline{a & b }, and \lstinline{a - b} respectively represent
the union, intersection, and difference of two sets \lstinline{a} and \lstinline{b}.

Following the vehicle selection,  a connection timeout of $2$ minutes is set for the UUVs using
\lstinline{setConnectionTimeout} (line~\ref{c:conf:timeout}).
Up to this time frame, the Dolphin runtime will assume the state of each UUV to be steady
in the event that the connection to it is (intermittently) lost.
The timeout setting attends to the fact that UUVs may operate underwater for a long time, during
which they may only report state through error-prone and intermittent
acoustic communications.

Each UUV task results from the \lstinline{UUVTask} Groovy function, parameterised by an argument~$i$;
since Dolphin is embedded in Groovy, we may integrate Groovy code within the program at will, making use
of standard imperative and object-oriented programming features.
The \lstinline{UUVTask} function yields the sequential composition of three tasks, as specified by the use of the \lstinline{>>} Dolphin-specific operator: (1) the actual survey task \lstinline{survey}$i$; (2) a notification action signalling readiness for rendez-vous, \lstinline{post ready:}$i$, that proceeds instantaneously and involves no vehicle interaction; and; (3) at the end, a ``station-keeping'' \lstinline{sk}$i$ task to make the UUV maintain a fixed position for rendez-vous.
The survey and station-keeping tasks are IMC plans (discussed in Section~\ref{sec:lsts}) 
to be executed by vehicles. Note that the survey task are merely identified by name, thus they are assumed to be pre-programmed  for the vehicles, whilst the station-keeping task is programmed inline in the code through the use of an DSL for IMC plans integrated into the Dolphin engine (see Section~\ref{sec:lsts:imc-dsl}).

Actual execution of tasks proceeds using an \lstinline{execute} block. In the code, we
see that the \lstinline{UUVs} are tasked with a composition of three 
tasks to execute concurrently, as specified by the use of the \lstinline{|} Dolphin operator:
\lstinline{UUVTask(1) | UUVTask(2)  | UUVTask(3)}. Also concurrently, \lstinline{UUV} 
is tasked with an \lstinline{allOf} event-based task that works as follows:
as each UUV task posts a \lstinline{ready:}$i$ notification, the \lstinline{allOf} task may consume it
in line with guard conditions \lstinline|when { consume ready: |$i$\lstinline|}| and 
executing a corresponding \lstinline{then} block, issuing an IMC plan \lstinline{rv}$i$ for rendez-vous. An \lstinline{allOf} task terminates  only when each of the components \lstinline{when}-\lstinline{then} blocks
have been completed.  In alternative to \lstinline{allOf}, a \lstinline{oneOf} task, specified with a similar structure, would require only one of the \lstinline{when}-\lstinline{then} blocks to fire, i.e.,
only one rendez-vous would execute instead.

The entire \lstinline{execute} block terminates when all component tasks terminate. This implies
that some vehicles may remain idle (executing some vehicle-dependent fallback behavior such as loitering) while waiting for the completion of ongoing tasks.
In sequence,  at the end of the program, the \lstinline{release UUVs + UAV} instruction decouples the vehicles
from the program, i.e., \lstinline{release} is the inverse operation of \lstinline{pick}. The instruction is redundant in this case, as an implicit release instruction is issued for all bound vehicles at the end of a program, but we show it in the example for the sake of clarity. Note 
that \lstinline{pick} and \lstinline{release} can generally be used at any point in a program to acquire and release vehicles on-the-fly, as illustrated later on (Fig.~\ref{fig:program2}).

\subsection{Task definition}
The full definition of Dolphin tasks is summarised by the BNF-style grammar of Fig.~\ref{fig:grammar}.
In addition to the task operators discussed above, a few others are defined, as follows:

\begin{figure}[h!]
\centering
{\begin{lstlisting}[language=dolphin,basicstyle=\scriptsize\bfseries\ttfamily, 
commentstyle=\scriptsize\bfseries\ttfamily\color{red}\bfseries,numberstyle=\scriptsize\ttfamily,keywordstyle=\scriptsize\bfseries\color{blue}\ttfamily,
morekeywords={imcPlan}]
Task := PlatformTask            // Platform task
     | action '{' Code '}'      // Program-level action
     | condition '{' Cond '}'   // Program-level condition
     | Task '>>' Task           // Sequential composition
     | Task '|' Task            // Concurrent composition
     | Task '[' VSet ']'        // Vehicle set allocation
     | allOf '{' WhenThen+ '}'  // All-of block
     | oneOf '{' WhenThen+ '}'  // One-of block (choice)
     | waitFor '{' Cond '}'     // Execution subject 
       then Task                //   to start condition
     | until '{' Cond '}'       // Execution subject  
       run Task                 //   to stop condition
     | idle Time                // Idle task
     | during Time              // Execution subject 
       run Task                 //   to time limit
     | watch Task               // Error handling
       onError '{' Code '}'
WhenThen := when '{' Cond '}' then Task                  
// For IMC-based platform
PlatformTask := imcPlan '(' Id ')' 
             |  imcPlan '{' IMC_DSL_Spec '}'
\end{lstlisting}}
\caption{Grammar for Dolphin tasks.\label{fig:grammar}}
\vspace{-0.4cm}
\end{figure}

\noindent --- The task allocation operator, \lstinline{T [ V ]}, defines
the allocation of task \lstinline{T} to vehicle set \lstinline{V}, and is already
implicit in the example program:  \lstinline{execute V1: T1, v2: T2, ...} (as in the program) is merely syntactic sugar for \lstinline{execute T1[V1] | T2[V2] | ... }. 

\noindent --- \lstinline{condition C} prevents progress until condition \lstinline{C} is satisfied.

\noindent --- Two operators combine event and control flow, waiting for an condition \lstinline{C} before starting or stopping a task \lstinline{T}, i.e., resp.\ \lstinline{waitFor C then T } and \lstinline{until C run T}.

\noindent --- Two other operators relate to behavior based on a  duration of time \lstinline{t}: idle tasks, \lstinline{idle t}, or task execution subject to a duration of $t$, \lstinline{during t run T}.

\noindent --- Finally, \lstinline|watch T onError { C }| watches for errors during the execution of \lstinline{T} (e.g., connection timeouts, internal errors), and executes \lstinline{C} if one is detected. The code in \lstinline{C} may include three special actions: \lstinline{ignore()}, \lstinline{propagate()}, and \lstinline{halt()} that respectively ignore the error, propagate it allowing the possibility of being handled by an higher-level \lstinline{onError} block, or halt the program immediately. 
A \lstinline!onError { propagate() }! behavior is the default for each task. When no \lstinline{onError} block is set to handle an error, the engine halts the program.

\begin{figure}[h!]
\vspace{-0.3cm}
{\begin{lstlisting}[xleftmargin=6em,language=dolphin,basicstyle=\scriptsize\bfseries\ttfamily, 
commentstyle=\color{red}\bfseries,numberstyle=\scriptsize\ttfamily,keywordstyle=\scriptsize\bfseries\color{blue}\ttfamily]
...
supplier = getSupplier();
while (supplier.active()) {
  // Await for task
  T = suppler.queue.await() 
  // Pick AUV 
  UAV = pick { type: 'UAV' } 
  // Execute
  execute UAV: 
          watch 
            ( during 10.minutes run T )
          onError {
            err -> {
              message 'Error: ' + err
              ignore()
          }
  // Release UAV
  release UAV
}
\end{lstlisting}
}
\caption{Illustration of additional task composition operators.\label{fig:program2}}
\vspace{-0.1cm}
\end{figure}

Some of these additional operators are exemplified in the fragment of Fig.~\ref{fig:program2}.
that also provides an illustration of the possible tight integration of Dolphin with Groovy/Java.
In the code, a task supplier is assumed to be implemented externally,
possibly mediating interaction with a human user. The supplier yields tasks to be executed by an UAV iteratively, that the Groovy \lstinline{while} loop executes while the supplier is active. In each iteration a UAV is picked and executes a supplied task for no more than 10 minutes. The constraint is conceivable, e.g., due to battery restrictions. Note that we use one UAV at a time, but different ones may be used in distinct iterations, allowing for vehicle churn in the network. An \lstinline{onError} block is set notifying the user of any errors with a message, but ignoring it otherwise, thus letting the program resume execution.

\subsection{The Dolphin engine}

The Dolphin engine is responsible for executing a program. We now provide some detail of how it 
works in abstract terms. The main state of a program may be modelled as
a partial map representing platform task allocations $\mathcal{A}: \mathcal{V} \mapsto \mathcal{T} \cup \{ \bot \}$, where $\mathcal{V}$ and $\mathcal{T}$ are resp.\ the vehicle and platform task domains 
and $\bot$ stands for no task allocation, plus a set $\mathcal{B} \subseteq \mathcal{V}$ 
of  vehicles that are bound (associated) to  the program.
This state varies dynamically according to the use of \lstinline{pick}, 
\lstinline{release}, and \lstinline{execute}:

\noindent ---
A \lstinline{pick} block first queries the platform for all connected vehicles 
$\mathcal{V}_c$, filters out those in $\mathcal{B}$ (already bound) and those that
do not match the selection filters (e.g., type, location, or payload as in the example program of Fig.~\ref{fig:program}),
obtaining a set $\mathcal{V}_f$. The result $\mathcal{V}_p$ of selected vehicles is a subset of $\mathcal{V}_f$ with a fixed cardinality specified by the \lstinline{count} parameter (in Fig.~\ref{fig:program}: 3 in  line~\ref{c:conf:pick:UUVs}, and 1 by default in line~\ref{c:conf:pick:UAV}) or the full $\mathcal{V}_f$ itself when the \lstinline{count} parameter is specified as a wildcard value (denoted \lstinline{ _ } in the language). The program state is then updated as follows:
$\mathcal{B} := \mathcal{B} \cup \mathcal{V}_p$, and $\mathcal{A}[v] := \bot, \forall v \in V_p$.

\noindent --- \lstinline{release }$\mathcal{V}_r$ (as in  Fig.~\ref{fig:program}, l.~\ref{c:release}) 
corresponds to updating the state as $\mathcal{B} := \mathcal{B} - \mathcal{V}_r$,
and removing any mappings for $\mathcal{V}_r$ in $\mathcal{A}$ (which should in any case
equal $\bot$ at this stage, i.e., no tasks will be executing for $\mathcal{V}_r$).

\noindent --- \lstinline{execute T} allocates vehicles on-the-fly to platform tasks encoded in \lstinline{T}, i.e., it dynamically changes $\mathcal{A}$, guided by explicit vehicle allocation specifications through the \lstinline{[ ]} operator. If no vehicle allocation is specified, an allowed alternative, the engine allocates all bound vehicles ($\mathcal{B}$).
Given that \lstinline{execute} only terminates only when the overall \lstinline{T} 
has fully completed, $\mathcal{A}[v]$ for a vehicle $v$ may alternate between a platform task $t$, while $t$ executes, and $\bot$, while $v$ is waiting for task allocation according to global flow of \lstinline{T}.
We stick to the informal description of previous sections for the overall flow of composed tasks in terms of sequence, concurrency, or event-flow, operating in the spirit of process calculi~\cite{ccs}. 
The engine basically
executes composed tasks by interpreting the abstract syntax tree of \lstinline{T} and the nature of each operator used in it. A full description is outside the scope of this paper for reasons of space.

In addition to the above characterisation of semantics, event-based program flow through notifications may be accomplished through  the \lstinline{post} and \lstinline{consume} operations (illustrated in Fig.~\ref{fig:program}), plus an additional \lstinline{poll} operation.
These use a simple tuple-space abstraction that can be modelled as multi-set $\mathcal{N}$ of key-value pairs. Each operation works as follows:

\noindent --- \lstinline{post k:v} adds \lstinline{(k,v)} to $\mathcal{N}$.

\noindent --- \lstinline{consume k:v} removes a \lstinline{(k,v)} pair from $\mathcal{N}$, if one
exists, and returns true in that case (note that the operation is normally used as a guard condition), otherwise it returns false. If \lstinline{v} equals the wildcard value \lstinline{_},  the operation tries to remove the oldest tuple in insertion order with key $k$.

\noindent --- \lstinline{poll k:v} merely inspects for the existence of a \lstinline{(k,v)} pair in $\mathcal{N}$, also possibly using the \lstinline{_} wildcard for \lstinline{v}.

\begin{figure}[!t]
{\begin{lstlisting}[language=Java,basicstyle=\scriptsize\bfseries\ttfamily,keywordstyle=\scriptsize\bfseries\ttfamily,
commentstyle=\scriptsize\color{red}\bfseries\ttfamily,escapechar=@]
public interface Platform ... {
  // Query nodes and tasks
  NodeSet getConnectedNodes();
  PlatformTask getPlatformTask(String id);
  // I/O
  String askForInput(String prompt);
  void displayMessage(String format, Object... args);
  // Extensibility 
  void 
   customizeGroovyCompilation(CompilerConfiguration cc);
  List<File> getExtensionFiles();
}
public interface Node { 
  // Attributes
  String getId();
  String getType();
  Position getPosition();
  Payload getPayload();
  // Task binding & vehicle release
  Task getRunningTask();
  void setRunningTask(Task task);
  void release();
  // Connection handling.
  double getConnectionTimeout();
  void setConnectionTimeout(double timeout); ...
}
public interface Task ... {
  String getId();
  TaskExecutor getExecutor();
  boolean allocate(NodeSet available, 
                   Map<Task,List<Node>> allocation);
}
public abstract class TaskExecutor ... { 
  // Lifecycle methods 
  protected abstract void 
    onInitialize(Map<Task,List<Node>> allocation);
  protected abstract void onStart();
  protected abstract CompletionState onStep();
  protected abstract void onCompletion(); ...
}
public abstract class PlatformTask implements Task {
  public abstract List<NodeFilter> getRequirements();
  public Optional<Position> getReferencePosition() ...
}
\end{lstlisting}
}
\caption{Java types for Dolphin platform bindings.}
\label{fig:bindings}
\vspace{-0.4cm}
\end{figure}
\subsection{Platform bindings}\label{sec:dolphin:bindings}

For using Dolphin with a concrete platform, a set of Java interfaces and abstract classes are defined for
implementation at the platform level. An overview of the main ones 
are listed in Fig.~\ref{fig:bindings}.
\lstinline{Platform} defines the abstract platform operations:
querying connected vehicles and existing platform
tasks, providing simple user I/O, and extensibility features that include the customisation
of Groovy compilation (e.g., automatic imports of certain APIs) and inclusion of
DSL extension files to load on startup.  
Vehicles are instances of \lstinline{Node}, with associated operations for querying basic attributes (id, type, payload, position), tasking them or releasing them from the program, and connectivity parametrisation. Tasks are instances of \lstinline{Task}, with an associated id, task allocation 
procedure, and task executor instance. A \lstinline{TaskExecutor} represents a task during 
execution, defining abstract methods for its lifecycle. Platform tasks are extensions
of tasks that must implement \lstinline{PlatformTask}, providing information regarding
vehicle requirements (e.g., vehicle type and payload), and an optional reference position
to aid task allocation by the Dolphin engine.

\section{Integration with the LSTS Toolchain}
\label{sec:lsts}

\subsection{Overview}

Over the years, LSTS developed several autonomous vehicles, with an associated
open-source toolchain~\cite{oceans13}.
This toolchain comprises three main components: 
(1) Neptus, a Java-based command-and-control tool for human operators to configure, plan, and monitor autonomous vehicles using a GUI; (2) DUNE, a C++ on-board software platform for autonomous vehicles, including a simulation mode, and; (3) IMC, an extensible message-based protocol for networked interoperability between all LSTS systems, with bindings in several languages such as C++ and Java.

A subset of IMC is dedicated to the specification, execution, and monitoring of tasks called IMC plans,
that we considered as the basic unit of computation for IMC-based Dolphin platforms.
An IMC plan is a sequence of maneuvers for a single vehicle, comprising
simple maneuvers such as waypoint tracking but also more complex ones such as
area surveys. Typically, IMC plans are programmed in Neptus and executed within vehicles by DUNE.

We developed two IMC-based Dolphin platforms taking form as (1) a simple
command-line based tool, and (2) integrated in the Neptus tool as a plugin.
For both, we had to implement the Dolphin abstract bindings, in particular vehicles 
and IMC plans respectively as instances of \lstinline{PlatformTask} and \lstinline{Node}, presented earlier in Section~\ref{sec:dolphin:bindings}. In complement, we developed a Groovy DSL specifically devoted to the specification of IMC plans, as we felt the need to define IMC plans directly within Dolphin programs, rather than just relying on the Neptus tool for that purpose.

%To interact with the vehicles we used the IMC communication protocol, having bindings in C++ and Java. The Java bindings for the protocol was used on the language core providing native support for LSTS systems. This protocol is also used to define and command plans under the form of \lstinline{PlanControl} messages, having an associated \lstinline{PlanSpecification} message attached to it. In the other hand, each vehicle receives and executes these types of plans through its onboard software DUNE (developed in C++).
%
%There was used two distinct LSTS software toolchain as platforms to implement the language: one running inside a plugin in the command and control software, Neptus. The other one, running in the terminal in a standalone version. Both versions have support to the language extensions which includes the generation of IMC plans inside a language script through the IMC DSL.

\vspace{-0.1cm}
\subsection{IMC standalone platform}
The IMC standalone platform is a command-line tool for executing Dolphin programs.
To support IMC interaction, we made use of the IMC Java bindings, and networking code
for discovering vehicles in the network using multicast UDP, 
and then exchanging messages with them over standard UDP. For IMC plan interaction, 
we made use of \lstinline{PlanSpecification} (for plan definition), \lstinline{PlanControl} (control), and \lstinline{PlanControlState} (monitoring) messages from the IMC specification\footnote{\url{http://github.com/LSTS/imc}}. The platform extends the Dolphin DSL with \lstinline{imcPlan} tasks,
identified by id or defined inline using the IMC DSL (discussed below).

\vspace{-0.1cm}
\subsection{Dolphin plugin for Neptus}
%Another implementation made, used the Neptus software as target platform, placing the language common runtime inside a console plug-in. which is the command and control unit developed in Java,

\begin{figure}[!t]
\centering
\includegraphics[width=0.85\columnwidth]{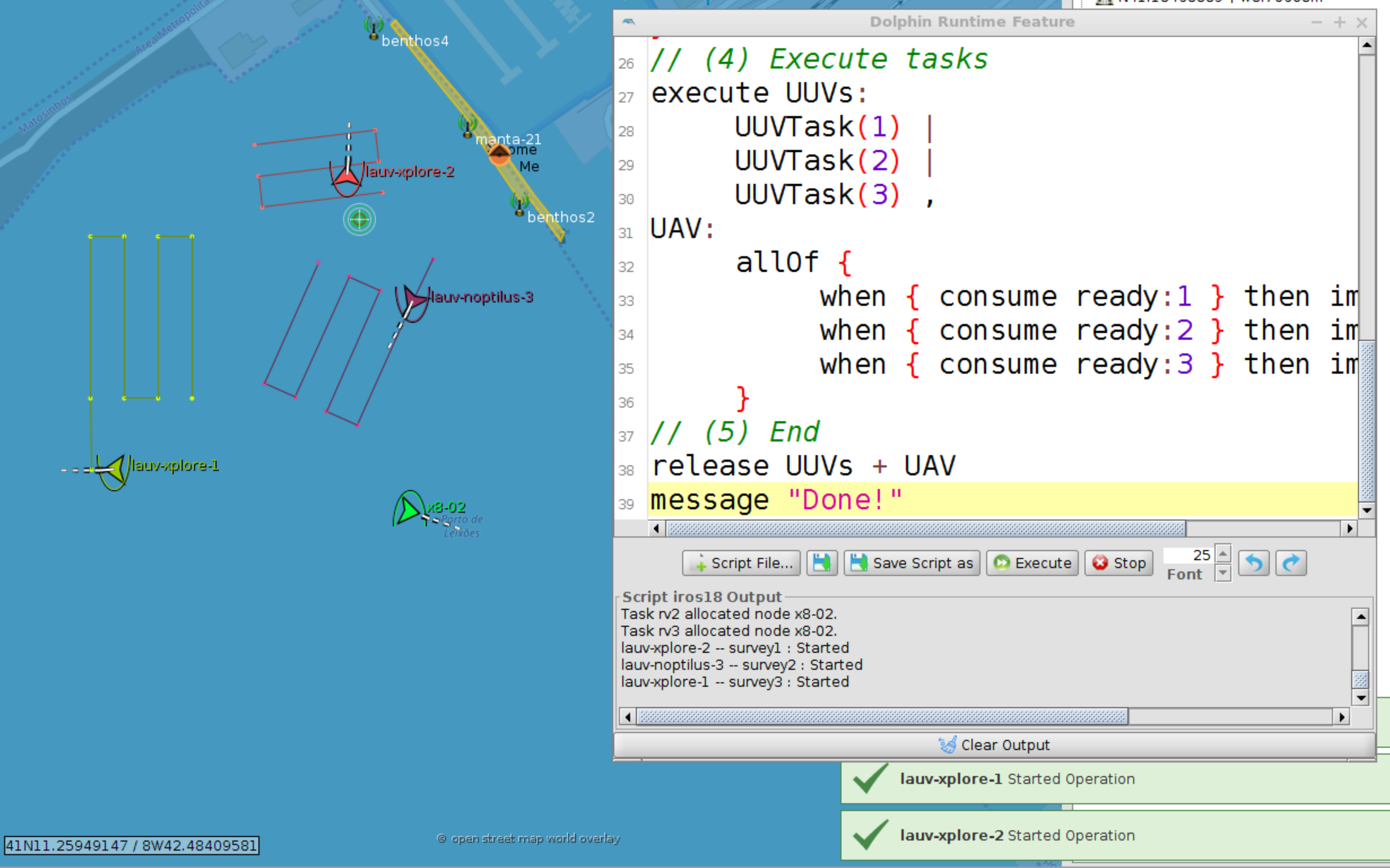}
\caption{Dolphin plugin running in Neptus.
\label{fig:plugin}}
\end{figure}

The Dolphin plugin
%\footnote{\url{https://github.com/LSTS/neptus/wiki/Creating-your-first-plug-in}} 
for Neptus allows users to edit and run Dolphin programs through a custom 
window, embedded in the overall GUI environment for editing and monitoring of IMC plans, %IMC plan edition and monitoring, 
as illustrated in Fig~\ref{fig:plugin}. During execution, the behavior of a program can be simultaneously
monitored using the Dolphin console and the standard Neptus GUI.   
The implementation traits are similar to the stand-alone IMC platform, 
apart from a delegation of networking functions to the pre-existing Neptus infrastructure, and 
an integration with the database of IMC plans associated to a Neptus console. 

%There is also support to the language extensions.

\vspace{-0.1cm}
\subsection{The IMC DSL}\label{sec:lsts:imc-dsl}

The IMC DSL
%\footnote{\url{http://github.com/DolphinDSL/imc-dsl}} 
may be used to automate IMC plan generation with relatively succinct textual descriptions. 
The listing of Fig.~\ref{fig:imc-dsl} illustrates a richer example than the one given in the sample program
of Fig.~\ref{fig:program}, defining an IMC plan with two maneuvers (\lstinline{Goto} and a \lstinline{Loiter}),
and associated parameterisation.

\begin{figure}[t!]
\centering
{\begin{lstlisting}[basicstyle=\bfseries\scriptsize\ttfamily,commentstyle=\color{red}\scriptsize\ttfamily]
t = imcPlan {
    // Id
    planName 'waterSurvey'
    // Set reference speed, depth, and location
    speed 1.5, Speed.Units.METERS_PS
    z     0.0, Z.Units.DEPTH
    locate Location.APDL
    // Goto maneuver, activating the camera payload
    move 30,-125
    goTo payload:[[name: 'Camera']]
    // Loiter maneuver
    move (-30,-50)
    loiter radius:100
}
\end{lstlisting}}
\caption{Task specified using the IMC DSL.\label{fig:imc-dsl}}
\vspace{-0.4cm}
\end{figure}

\vspace{-0.1cm}\section{Field tests}
\label{sec:experiments}

Dolphin has been evaluated in field tests that took place at the Leixões harbour, and in open sea at Tróia during the 2017 Rapid Environmental Picture (REP'17) exercise\footnote{\url{http://rep17.lsts.pt}} in collaboration with the Portuguese Navy~\cite{oceans18,keilaMSC}. Here we only present results for the example scenario/program of Section~\ref{sec:dolphin:example} at Leixões.

\subsection{Setup}
%Here we provide a summary of the first time the language was tested on a real operational scenario. These test consisted in a typical bathymetry mission, using vehicles equipped with sonar sensor to do the mapping inside an harbour. 
%It allowed the evaluation of the language behaviour outside simulation scope, enhancing possible improvements. %TODO mention timeout of 20secs -> vehicles didn't submerge-> became user-defined parameter
%%characterization

\begin{figure}[!t]
\centering
\subfloat[IMC plans in Neptus.]{
\includegraphics[width=0.75\columnwidth]{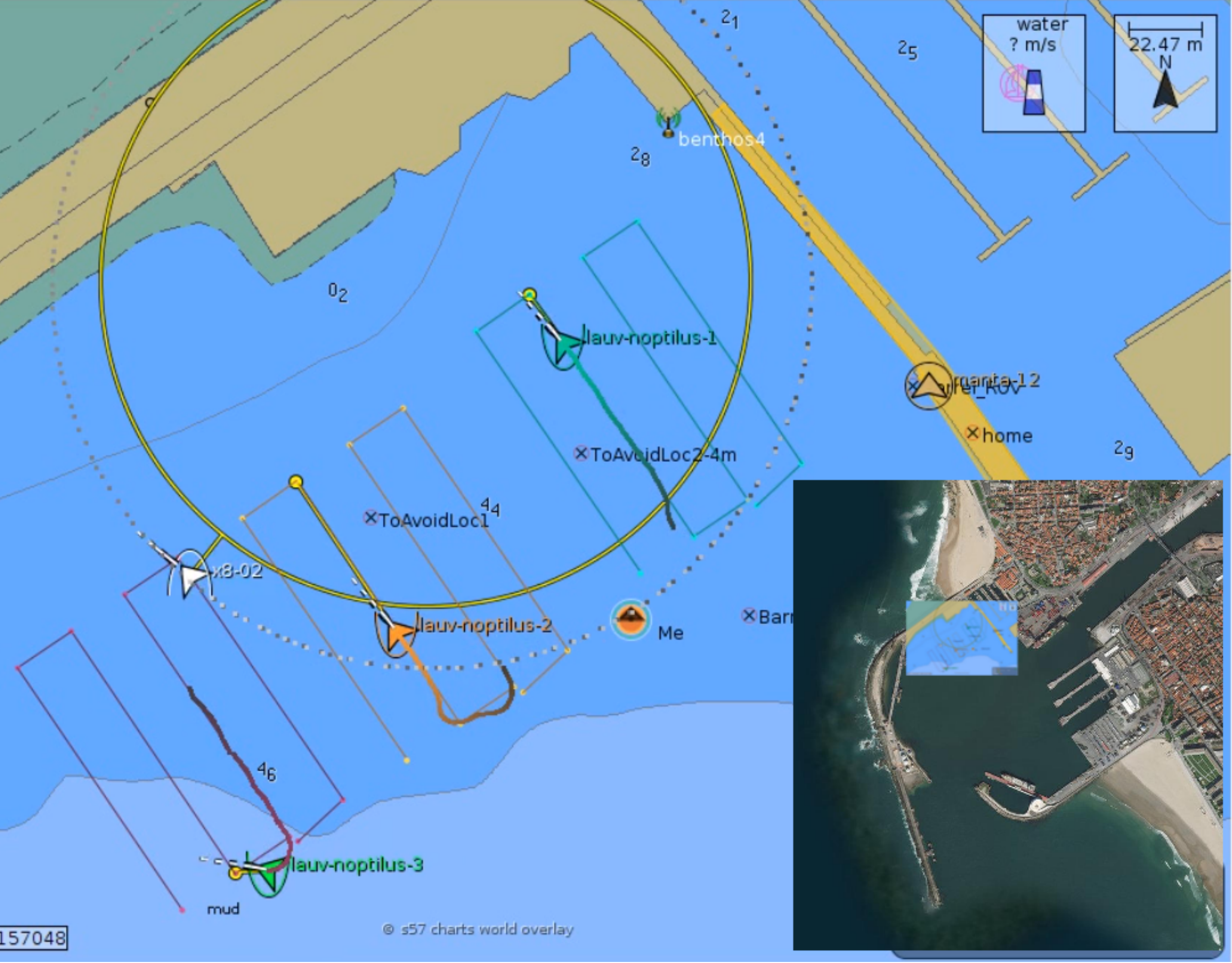}
 \label{fig:apdl:neptus}
 }
 \vspace{-0.2cm}
 
\subfloat[Noptilus UUVs (orange-colored vehicles).]{
  \includegraphics[width=0.75\columnwidth]{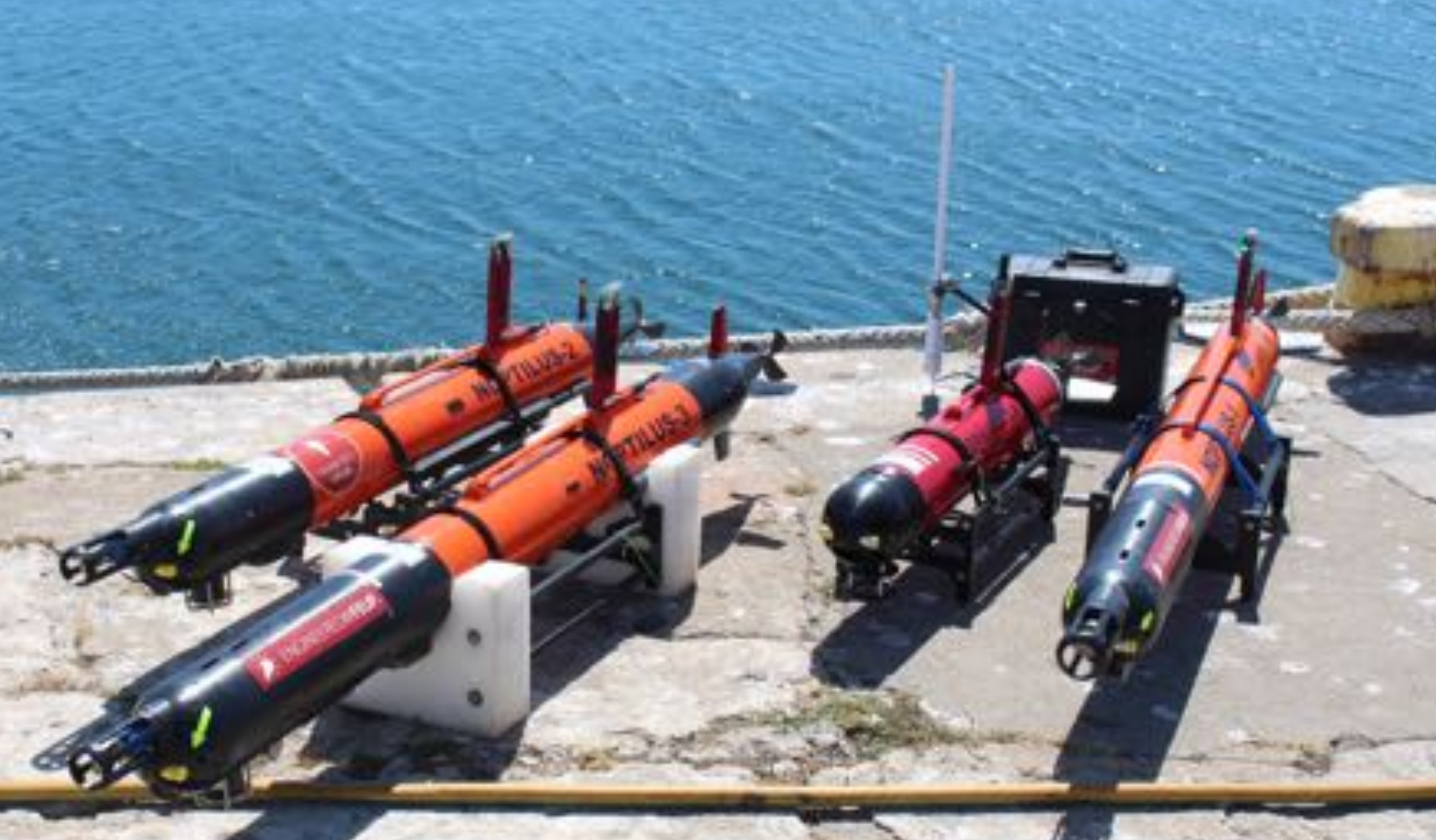}
  \label{fig:apdl:systems}
}
\caption{Test setup.\label{fig:apdl}}
\vspace{-0.5cm}
\end{figure}

In Fig.~\ref{fig:apdl} we present an overview of the scenario (\ref{fig:apdl:neptus}) and a photo of the vehicles we used (\ref{fig:apdl:systems}). The scenario overview is a Neptus screenshot, with the scenario already being executed, depicting three IMC plans for the water surveys to be executed by UUVs, and one
of the rendez-vous plans executed by a simulated UAV (the circular loiter).
Due to operational restrictions at Leixões, we could only deploy a simulated UAV as part of the control loop. 
The harbour location is shown bottom right in the same image\footnote{\url{https://goo.gl/maps/hHvdiTt3xAH2}}.
For the tests, we used three LAUV-class vehicles~\cite{lauv} shown in the photo of Fig.~\ref{fig:apdl:systems} (an additional fourth vehicle shown was used for unrelated operations), named Noptilus-1, Noptilus-2, and Noptilus-3. The same photo shows a Manta communications gateway~\cite{oceans13}, used for WiFi and underwater
communications between vehicles and Neptus consoles.

\begin{figure}[!t]
\centering
\subfloat[Execution timeline.]{
\includegraphics[width=0.80\columnwidth]{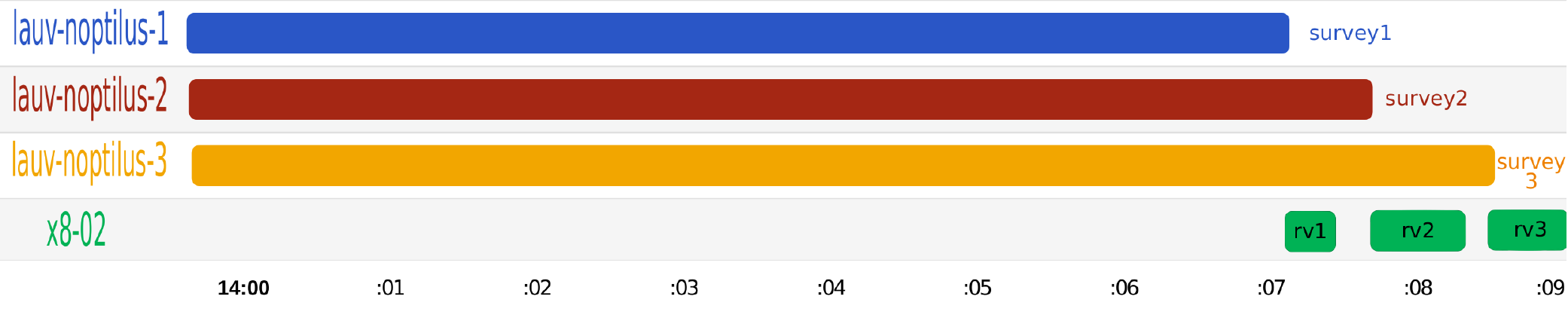}
\label{fig:results:timeline}
}
\vspace{-0.2cm}

\subfloat[Vehicle positions (XY).]{
\includegraphics[width=0.85\columnwidth]{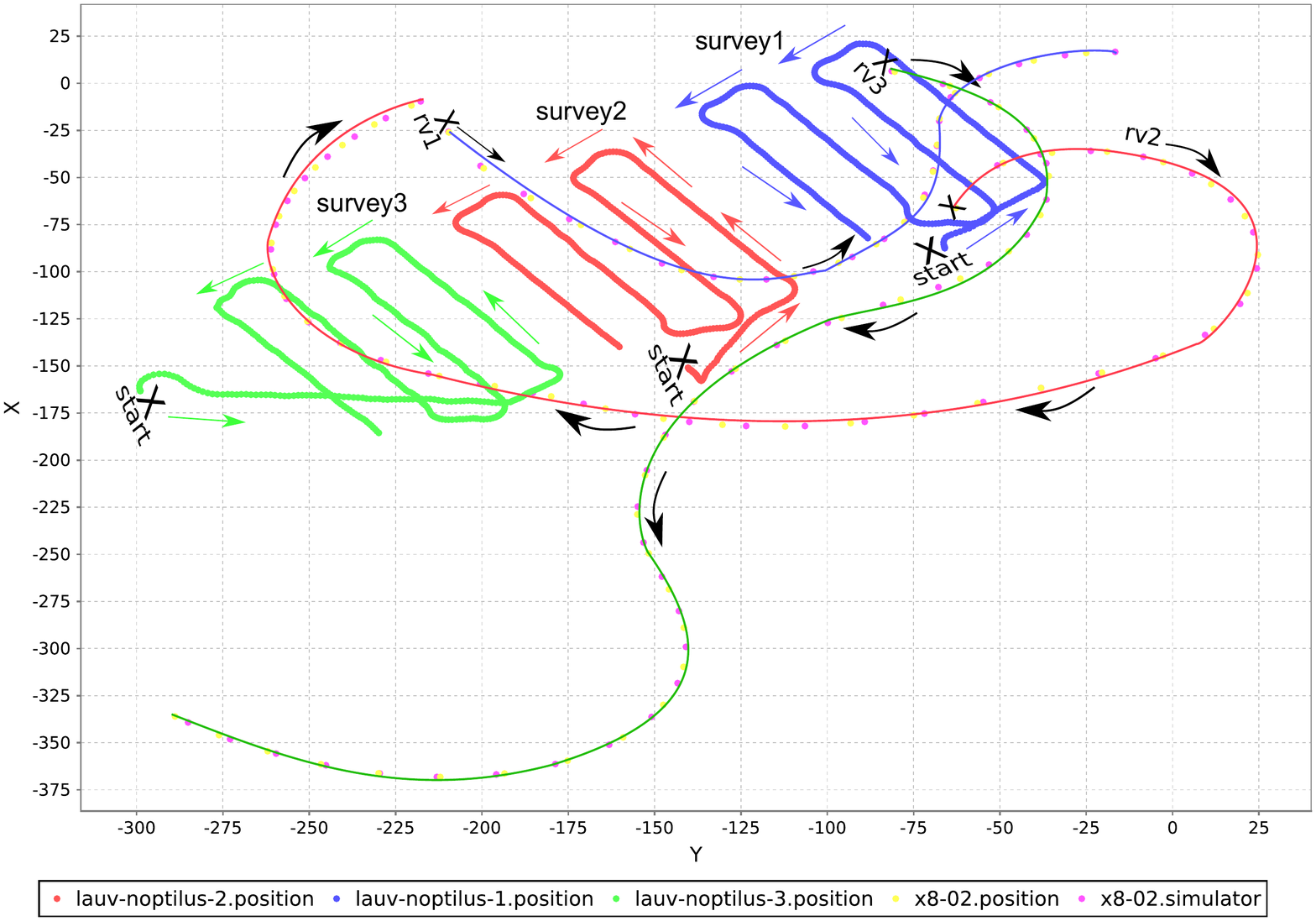}
\label{fig:results:xy}
}
\vspace{-0.2cm}

\subfloat[Bathymetry values measured using DVL.]{
\includegraphics[width=0.85\columnwidth]{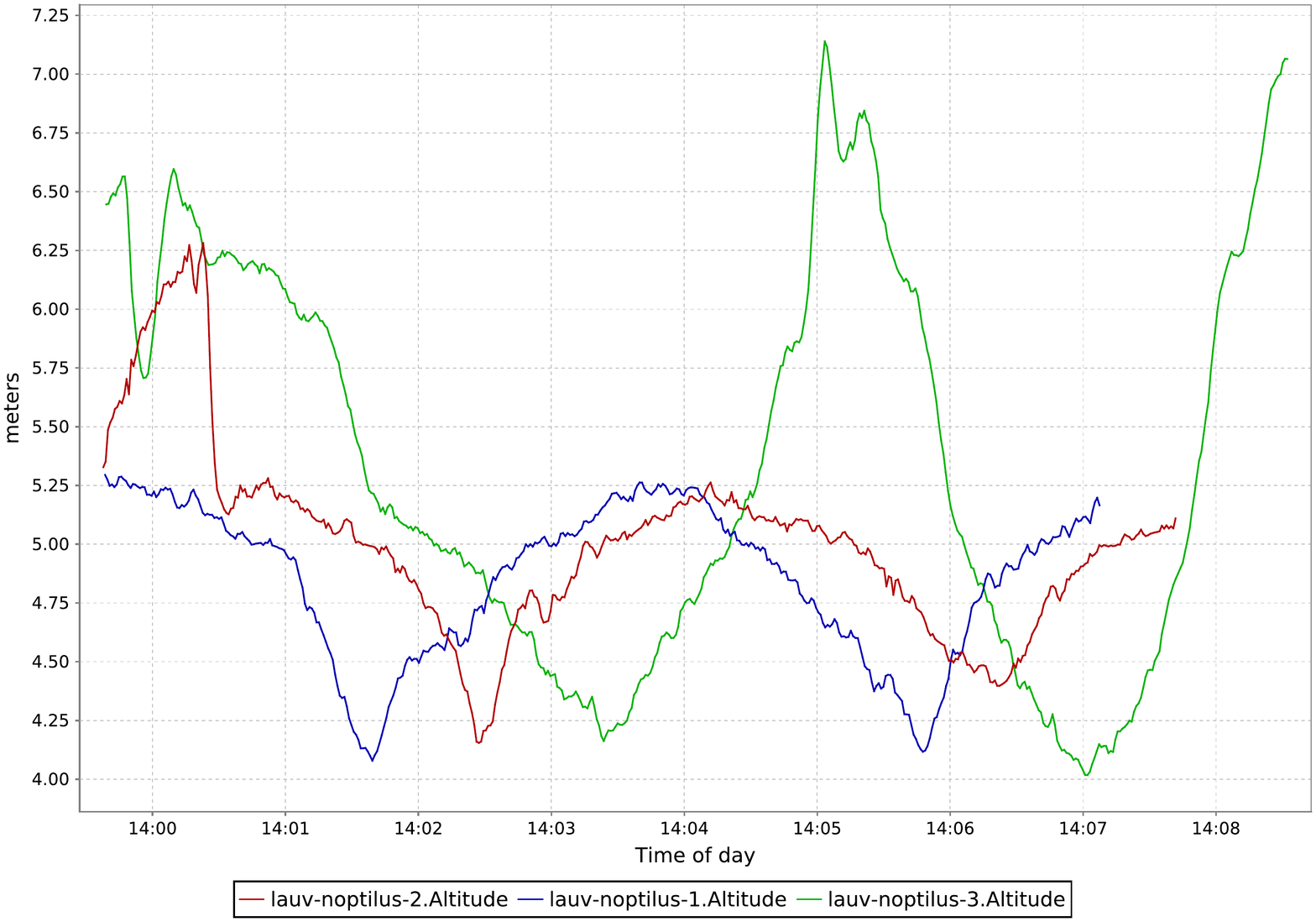}
\label{fig:results:z}
}
\vspace{-0.2cm}

\subfloat[Bathymetry map.]{
\includegraphics[width=0.85\columnwidth]{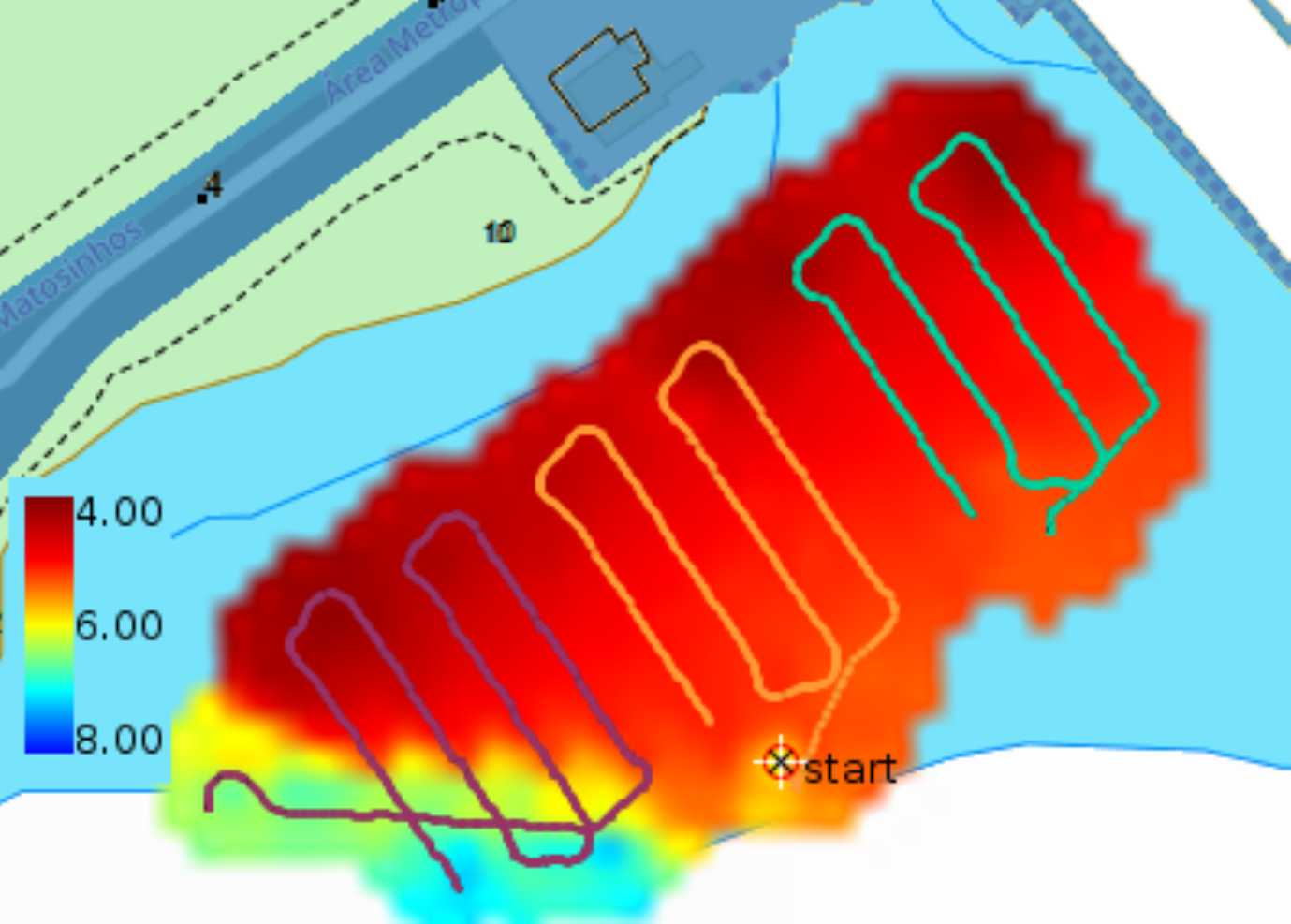}
\label{fig:results:colormap}
}
\caption{Results for the field test scenario.\label{fig:results}}
\vspace{-0.5cm}
\end{figure}

Each of the UUVs was equipped with a DVL, allowing us to fulfil the purpose of
gathering bathymetry data (distance to the seafloor) for the area of operation, thus
the vehicles were programmed to operate at the surface. 
A variant of the  program in Fig.~\ref{fig:program} was edited in Neptus using the Dolphin plugin, and the associated IMC plans were edited using a Neptus plan edition console. The main differences of the program variant were that we used named vehicle selection (through an \lstinline{id} attribute in \lstinline{pick} blocks) to avoid unexpected movements in the relatively short area we had for operation,
and that the IMC plans for UUVs did not include a station-keeping maneuver after the survey.
Moreover, we left the connection timeout for UUVs (through \lstinline{setConnectionTimeout}, discussed in Section~\ref{sec:dolphin}) to the default minimum of 1 minute, since they operated at the surface and always in WiFi reach. During tests in Tróia~\cite{oceans18,keilaMSC}, 
the same overall scenario was executed, but a timeout of several minutes was set instead to cope with connectivity issues raised by the use of underwater acoustic communications. The difference is that, in Tróia,
the UUVs operated underwater for long periods with a bottom-tracking  approach (i.e., maintaining a constant distance to the seafloor) for the purpose of gathering side-scan sonar data. 

\subsection{Results}

The results of one of the executions of the target scenario are shown in Fig.~\ref{fig:results:timeline}
comprising: a timeline of executed IMC plans (\ref{fig:results:timeline}), vehicle positions (\ref{fig:results:xy}), DVL measurements over time per vehicle (\ref{fig:results:z}), 
and a global bathymetry plot derived from the DVL measurements (\ref{fig:results:colormap}).

As shown in the timeline of Fig.~\ref{fig:results:timeline}, the whole execution took roughly 9 minutes.
The UUVs started their surveys simultaneously, but each of the three surveys terminated at different times.
Even if the surveys had similar path lengths, non-linear factors such as initial position, sea currents, vehicle calibration obviously impact on execution times. As each survey terminated, the UUV initiated a corresponding rendez-vous. Purely by chance, the order of completion of the survey and rendez-vous maneuvers was in line with vehicle numbering (1, 2, 3). Other executions of the scenario yielded a different order of completion.

The XY plot in Fig.~\ref{fig:results:xy} shows that the UUV paths are in line with the programmed survey plans. In the same figure, for clarity, the UUV path is annotated in terms of rendez-vous plan trajectories (\lstinline{rv1}, \lstinline{rv2}, and \lstinline{rv3}), 
where we can notice an approach of the UAV to each survey area.  
For a non-cluttered plot, we omit the UAV paths in between rendez-vous plans, during which the vehicle  loitered in the air. The DVL values of Fig.~\ref{fig:results:z} indicate different depths and variations according to each vehicle/survey area, with values ranging approximately from 4.0 to 7.5 meters.
The vehicle logs were collected and post-processed to obtain the bathymetry plot of the overall operation area in Fig.~\ref{fig:results:colormap}.

%This parameter became user-defined, assuming values from five seconds to one hour.

% including one for a variant of the rendez-vous program
%TODO presented earlier in Section TODO, for which we present results. 

\vspace{-0.2cm}\section{Related work}\label{sec:rwork}

Dolphin follows on from 
NVL~\cite{sac15}, also a task orchestration language.
Similarly to Dolphin, NVL defines primitives for selecting vehicles over a network, but only 
a single ``step'' primitive for firing  tasks concurrently for multiple vehicles
in contrast to compositionally-defined tasks in Dolphin.
NVL requires explicit task-vehicle allocation with the granularity 
of single vehicles rather than vehicle sets, and is not extensible out-of-the-box for new constructs
or robotic platforms,  as these require direct changes  
on the base code of NVL. Integration with the LSTS toolchain is also very limited:
apart from the use of IMC, a command line executor and a language-specific editor had to be used
with no interface to Neptus. It also relied on previously programmed IMC plans,
in contrast to the Dolphin integration where we have the option of using the IMC DSL.

We now survey other DSLs for coordinated task
execution of networked robotic systems, then make a final discussion 
contextualising Dolphin in the overall research landscape.

Karma~\cite{karma} implements an orchestration architecture for programming 
 micro-aerial vehicle (MAVs) networks, called the hive-drone model.
Orchestrated behavior is conducted by a centralised coordinator,
called the hive. Tasks, called drone behaviours, can be allocated to multiple
vehicles and interact through a centralised datastore running at the hive.
Each behavior is specified independently, with an
associated activation predicate and a progress function, both of which 
feed on information the hive datastore. The progress function governs the on-the-fly allocation 
of more or less drones by the hive to a single behavior, according to task completion
results reported by drones. Though there are no composition constructs,
composed behaviours may be defined implicitly by datastore value dependencies. 
The hive-drone model of Karma is also used by the Simbeeotic simulator 
for MAV swarms~\cite{simbeeotic}. 
Other languages used with aerial vehicles follow the spirit of Karma's centralised architecture,
using different kinds of abstractions, e.g., TeCoLa~\cite{tecola} is 
a Python DSL built around the notions of vehicle teams and services accessible via remote procedure
calls, and CSL~\cite{csl} uses reconfigurable Petri nets for task orchestration.

Proto~\cite{proto} is a functional programming language for homogeneous
robots. The conceptual approach is based on computational fields,
whereby a collection of devices approximates a continuous field in space/time.
A Proto program specifies choreographed swarm-like behavior through the composition of operators for restricting execution in space and time, feedback-loops that define state and execution flow,  and neighbourhood-based computation. Programs are compiled to abstract bytecode that is deployed using a viral propagation mechanism over the network, and then executed by each robot in distributed manner.
Bytecode execution uses a stack-based virtual machine for programs that may run on very lightweight microprocessor chips with only a few KB of RAM. Protelis~\cite{protelis} is a more recent language based on Proto, embedded in/interoperable with Java.

Meld~\cite{meld} is a logic programming language that, like Proto, also realizes a top-down synthesis approach. 
The state at each robot defined by a set of logical facts that evolve according to rules producing new facts. 
Rules take into account local robot state, but also a special constructs that access the state of neighbouring robots. Like Proto, the Meld compiler derives local robot programs that run on very lightweight
embedded platforms. 

Buzz~\cite{buzz} is a DSL for programming heterogeneous robot swarms.
Programs directly run locally on each robot, instead of being derived 
from top-level specification as in Proto or Meld, but have a notion of belonging
to a specific swarm with the intent of emergent collective behavior. 
Multiple swarms of heterogenous robots may be formed, making use of
 primitives for swarm formation, neighbourhood-based
queries and broadcast operations, and additional communication
through a distributed tuple space.
Buzz is compiled onto abstract bytecode that is executed by a
lightweight virtual machine, and can be integrated with C/C++ code.
In particular, ROSBuzz~\cite{rosbuzz} integrates Buzz in ROS. 
In the line of Buzz, Swarmorph-script~\cite{swarmorph} is a rule-based language 
for self-assembling robot swarms for morphogenesis. 

Voltron~\cite{voltron} is a language for mobile sensing using autonomous vehicles.
%Like Proto, the language decouples the definition of tasks from their actual realisation using
%vehicles. 
The network of available vehicles is tasked as a whole, regardless
of how many vehicles are available and without having to associate tasks to vehicles.
Tasks are defined by actions to be executed at a set of locations, using a key-value store for
coordinated behavior, and can be started/stopped
and engage less/more vehicles on-the-fly, in line with an active sensing strategy that accounts for the evolution of accomplished goals. Voltron is implemented through source-to-source translators
to C++ and Java, and supports centralised and distributed execution modes. In the distributed
execution mode, every vehicle bound to the same task executes the same program (as in Proto), 
and virtual synchrony  mechanisms are employed for the consistency of the shared key-value store.

Summarising the above discussion, we observe three main types of approach for globally tasking
autonomous vehicle networks:
(1) orchestration, where a single program/coordinator tasks nodes on-the-fly without need for explicit coupling between nodes and optional node-to-node communication (Dolphin, NVL, Karma, TeCoLa, CSL); (2) choreography, where nodes are programmed through a global specification (Proto, Protelis, Meld) leading to synthesised programs that organise as swarms, and; (3) distributed programs without an explicit global specification but emergent swarm behavior (Buzz, Swarmorph-script). Voltron allows both orchestration and choreography, given the choice between centralised and distributed implementations.

Dolphin shares common features to the discussed languages, such as: the ability of tasking vehicle teams in most of them; a design for extensibility and integration with other languages as in the case of TeCoLa, Buzz, Protelis, or Voltron, and; an explicitly compositional definition of tasks as in Proto and Meld. In regard to the later aspect, a distinguishing feature of Dolphin lies in the use of a process-calculi approach for task definition. On the other hand, Dolphin lacks relevant features, some of which discussed as future work in the next section:  tasks that are associated dynamically to multiple vehicles in line with a notion of progress as in Karma and Voltron, or the possibility of neighbour-based and team-level primitives for cooperative behavior as in Meld, Proto, Buzz, and Voltron.

\section{Conclusion}\label{sec:conclusion}

We presented Dolphin, a programming language for task orchestration in autonomous vehicle networks,
its integration with the LSTS toolchain for autonomous vehicles, and the use of the language in a field tests involving multiple vehicles. A Dolphin program is a global 
specifications of multi-vehicle tasks that are defined compositionally.
The language is also naturally extensible by virtue of its definition as a Groovy DSL and of 
the abstract platform bindings that made the LSTS toolchain integration possible.

As future work, we are interested in extending Dolphin in a number of ways such as:
the representation of human operators or sensors as nodes, in addition to vehicles;
vehicle interaction constructs in support of cooperative tasks, in
complement to the centralised tuple-space scheme we now employ;
tasks that aggregate vehicle sets with varying cardinality, for example according
to progress measured as set of accomplished goals or in reaction to vehicle faults during
execution, and; more expressive operators for a space-time characterisation of task flow, 
for now only implicit in vehicle selection criteria or the nature of IMC-based plans we used
in the LSTS platform. Additional platform bindings for popular toolkits such as ROS~\cite{ros} would also be interesting, beyond the current support for the LSTS toolchain and the work in progress regarding MAVLink~\cite{dolphin-site,mavlink}. 
%Another line of work concerns a quantitative verification back-end that examines a Dolphin program and static platform definitions (e.g., vehicle capabilities, task durations) to establish feasibility properties, e.g., as maximum time-to-completion or the minimum number of vehicles required for program execution.

%\vspace{-0.2cm}
\bibliographystyle{IEEEtran}
\bibliography{refs}

\noindent{\scriptsize{\bf Acknowledgements}
This work was partially funded by European Commission, ERDF (P2020), under the projects Endurance (NORTE-01-0247-FEDER- 017804) 
and SMILES/TEC4GROWTH (NORTE-01-0145-FEDER-000020).}

\end{document}